%% file: main.tex
\documentclass{article}
\pdfoutput=1
\usepackage{graphicx}
\usepackage[final]{ap_2017}
\usepackage[dvipsnames]{xcolor}

\usepackage[utf8]{inputenc} 
\usepackage[T1]{fontenc}    
\usepackage{hyperref}       
\usepackage{url}            
\usepackage{booktabs}       
\usepackage{amsfonts}       
\usepackage{nicefrac}       
\usepackage{microtype}      
\usepackage{multirow}
\usepackage{soul}
\usepackage[dvipsnames]{xcolor}
\usepackage{array}
\newcolumntype{P}[1]{>{\centering\arraybackslash}p{#1}}
\newcolumntype{M}[1]{>{\centering\arraybackslash}m{#1}}

\newcommand{\taskbot}{\textbf{TacoBot}}

\newcommand{\nop}[1]{}

\title{Bootstrapping a User-Centered Task-Oriented Dialogue System}

\author{
    Shijie Chen\thanks{Team Leads. Equal Contribution. } , Ziru Chen\footnotemark[\value{footnote}] , Xiang Deng\thanks{Team members in alphabetical order.} , Ashley Lewis\footnotemark[\value{footnote}] , Lingbo Mo\footnotemark[\value{footnote}] , Samuel Stevens\footnotemark[\value{footnote}] \\ 
    \textbf{Zhen Wang\footnotemark[\value{footnote}] , Xiang Yue\footnotemark[\value{footnote}] , Tianshu Zhang\footnotemark[\value{footnote}] , Yu Su\thanks{Faculty advisors.} , Huan Sun\footnotemark[\value{footnote}]}\\	
    \\
    The Ohio State University\\
    \\
	\texttt{\{chen.10216, chen.8336, deng.595, lewis.2799, mo.169, stevens.994,} \\
	\texttt{wang.9215, yue.149, zhang.11535, su.809, sun.397\}@osu.edu}
}

\begin{document}

\maketitle

\begin{abstract}
We present \taskbot, a task-oriented dialogue system built for the inaugural Alexa Prize TaskBot Challenge, which assists users in completing multi-step cooking and home improvement tasks. TacoBot is designed with a user-centered principle and aspires to deliver a collaborative and accessible dialogue experience. Towards that end, it is equipped with accurate language understanding, flexible dialogue management, and engaging response generation. Furthermore, TacoBot is backed by a strong search engine and an automated end-to-end test suite. In bootstrapping the development of TacoBot, we explore a series of data augmentation strategies to train advanced neural language processing models and continuously improve the dialogue experience with collected real conversations. At the end of the semifinals, TacoBot achieved an average rating of $3.55/5.0$. 
\end{abstract}

\input{intro-v2}
\input{overview}
\input{NLU}
\input{DM}
\input{TaskSearchRerank}
\input{NLG}
\input{UE_short}
\input{test-suite}
\input{results}

\input{conclu-v2}

\subsubsection*{Acknowledgments}
We would like to thank Amazon for financial and technical support as well as colleagues in the OSU NLP group for their valuable comments. We specially thank Yu Gu for his contribution that helped take this off the ground. 

\clearpage
\newpage

\bibliographystyle{acl_natbib}
\bibliography{main}

\end{document}

%% file: intro-v2.tex
\section{Introduction}

This paper presents \taskbot, a dialogue system built for the first Alexa Prize TaskBot Challenge, which assists users in completing multi-step cooking and home improvement (DIY) tasks. We envision TacoBot to be \textit{user-centered} and deliver a collaborative and accessible conversational experience. To this end, TacoBot must demonstrate superior capabilities, including accurate language understanding, flexible dialogue management, and engaging response generation. 

We face several challenges in achieving the vision: (1) \textit{lack of in-domain training data,} especially given modern neural models are data-hungry and crowdsourcing large-scale annotations is costly, (2) \textit{domain shift of language patterns} from the general domain (on which existing models are trained) to the cooking and DIY domain (the foci of this competition), and 
(3) \textit{the noisy nature of real user utterances,} which contain long-tail patterns that are difficult for models to understand.  


To conquer these challenges, our endeavors include: (1) We explore a series of data augmentation strategies, including leveraging GPT-3 \citep{brown2020language} to synthesize large-scale training data, which enable us to substantially improve natural language understanding and search and lay a firm foundation for the entire TacoBot system. (2) We annotate real user conversations to build evaluation datasets for most modules. Compared to simulated data, these realistic datasets better reflect the authentic distribution of user inputs and reveal fatal deficiencies of our models. (3) We translate observed user behaviors into actionable design guidelines, based on which we continuously improve our dialogue management module to be more flexible and our generated responses more engaging.  
(4) We further build an automated end-to-end test suite from scratch, which allows us to identify issues in our system efficiently and fix them before deployment. As a result of all these efforts, among the rated conversations in the semifinals, TacoBot achieved an average rating of $3.55/5.0$ and a task completion rate of $20.08\%$.

%% file: overview.tex
\begin{figure}[t]
  \centering
  \includegraphics[width=\columnwidth]{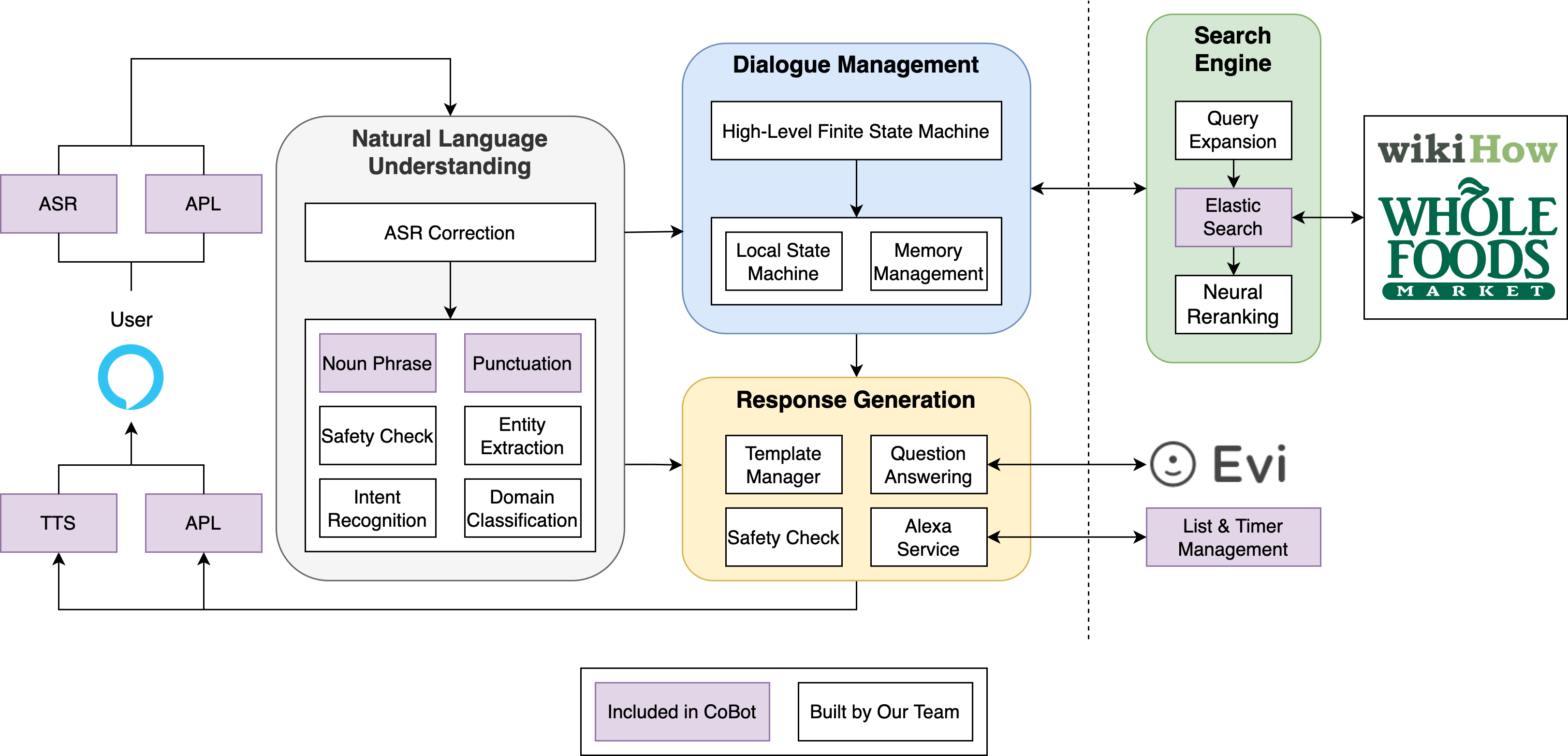}
  \caption{Overall System Design.}
  \label{fig1}
\end{figure}
\section{System Overview}
\label{overview}
Our dialogue system (Figure \ref{fig1}) is built on top of the CoBot framework \citep{abs-1812-10757}. At each turn, a user can interact with our system by speaking or using the touch screen (if available). In the speaking mode, Alexa’s Automatic Speech Recognition (ASR) transcribes the user's speech into a text utterance. In the touching mode, the touch event is described as a set of argument-value pairs in Alexa Presentation Language (APL). 


Upon receiving the user input, our system first executes the natural language understanding (NLU) pipeline (Section \ref{nlu}), which first corrects potential ASR errors and then fetches all other modules, hosted as remote EC2 instances. 
To minimize latency, we use CoBot to run all modules in parallel. 

When the NLU pipeline returns results successfully, our system calls the dialogue management (DM) module (Section \ref{dm}). It relies on a hierarchical finite state machine to control the dialogue flow, handle exceptions, and advance the conversation towards task completion implicitly. Meanwhile, it manages our system's dialogue context in DynamoDB and stores the results of our search engine (Section \ref{search}) if detecting a task request. Lastly, DM decides on which response generation modules to execute.

Then, our system runs the selected response generation modules (Section \ref{nlg}) concurrently. Most of the modules are template-based, and we additionally develop a neural question answering module to address user questions. The final textual or multimodal response is rendered by Alexa's SSML Text-To-Speech (TTS) and APL services before delivered to the user. 

%% file: NLU.tex
\section{Natural Language Understanding}
\label{nlu}

\subsection{TacoBot NLU Pipeline}
The NLU pipeline runs at the beginning of each dialogue turn. 
We blend powerful pre-trained language models and reliable rule-based approaches to build four NLU components: (1) ASR Error Correction, (2) Intent Recognition, (3) Task Name Extraction, and (4) Task Domain Classification. 


\textbf{ASR Error Correction.} Per our analysis of TacoBot conversations, ASR error is a major cause of user frustration, especially when users try to make a choice or give a command. For example, the ASR transcription of ``step four" often becomes ``step for," hindering our system from recognizing this navigation intent. To mitigate this problem, we annotate some common ASR errors in certain dialogue states and correct them by string matching. 
\textbf{Hierarchical Intent Recognition.} To support diverse user initiatives, we define 11 dialogue intents under four categories: 
\begin{itemize}
    \item \textbf{Sentiment.} On each turn, the user may confirm or reject the bot's response. Hence, we have three intents, \texttt{Affirm}, \texttt{Negate} and \texttt{Neutral}, to identify the polarity of the user utterance.
    
    \item \textbf{Commands.} The user can drive the conversation with several commands: \textbf{Task Request}, 
    \textbf{Navigation} (\texttt{More/Less Choice} in Task Catalog to view candidate tasks; \texttt{Forward (X Steps)}, \texttt{Backward (X Steps)}, and \texttt{Go To Step X} in Task Execution to walk through the steps), \textbf{Detail Request}, and \textbf{Task Complete}. The user can also terminate the ongoing conversation with \textbf{Stop} intent at any time.
    
    \item \textbf{Utilities.} Besides two common utility intents, \textbf{Repeat} and \textbf{Help}, we have a \textbf{Question} intent to capture various user questions throughout the conversation. Also, we support the Alexa list and timer management features using separate intents \textbf{List} (\texttt{Add} or \texttt{Remove} items) and \textbf{Timer} (\texttt{Set}, \texttt{Pause}, \texttt{Resume}, or \texttt{Cancel} the timer). 
    
    \item \textbf{Exception.} To avoid changing dialogue states by mistake, we have one additional intent for out-of-domain inputs, such as incomplete utterances and greetings.
\end{itemize}
Moreover, real user initiatives are more complex and may include multiple intents at a time. 
For instance,  \textit{"No, I want to know how to wash my car."} should be classified as both \textbf{Sentiment (\texttt{Negate})} and \textbf{Task Request} intents.
Therefore, we formulate intent recognition as a multi-label classification problem and filter model predictions by dialogue states. Certain intents, such as \textbf{Navigation}, are further parsed into fine-grained intents by regular expressions.

\begin{figure}[h]
  \centering
  \includegraphics[width=0.7\columnwidth]{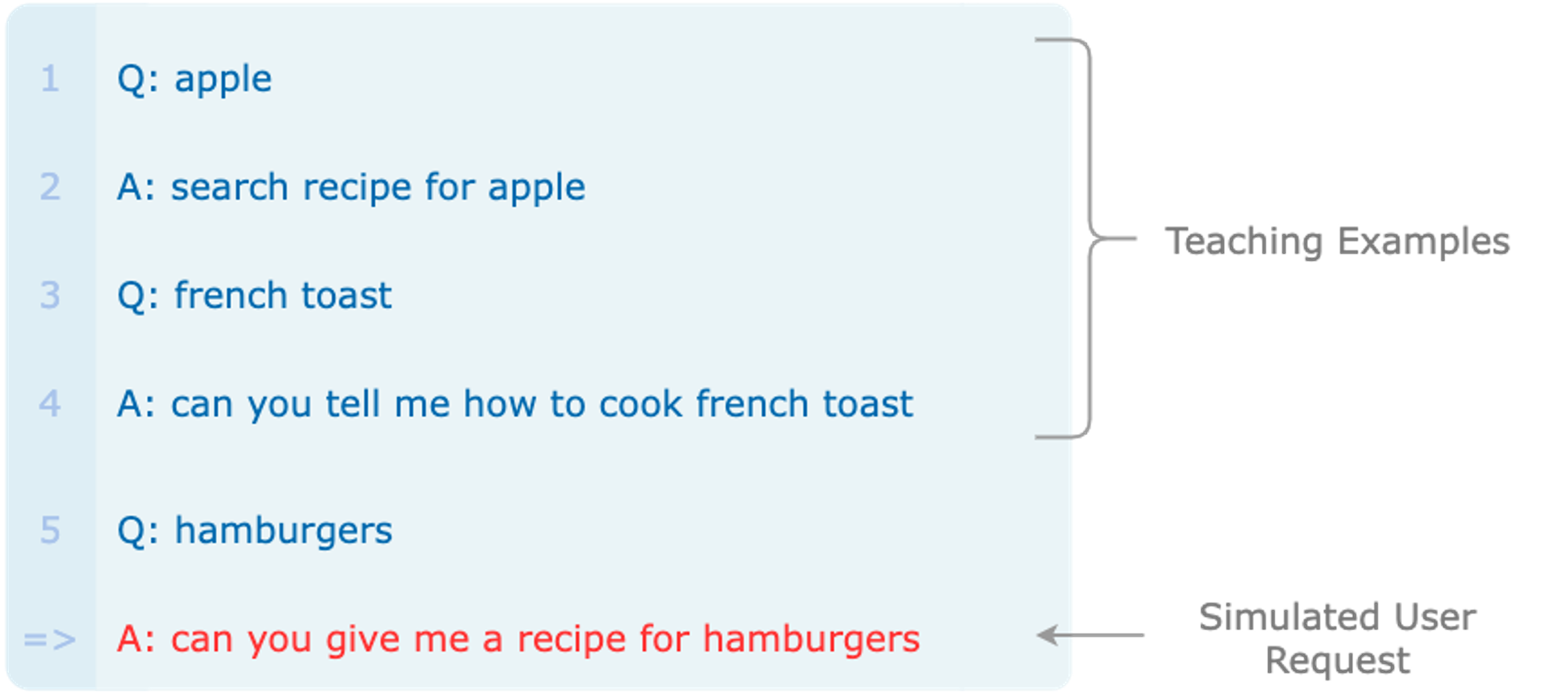}
  \caption{Using GPT-3 to Simulate Task Requests.}
  \label{fig2}
\end{figure}

To develop this multi-label classification model, we use data augmentation and domain adaptation techniques to get high-quality training data. We first reuse existing datasets~\citep{dstc8} for common conversational intents such as \textbf{Sentiment} and \textbf{Question}. For other intents, we leverage the in-context learning ability of GPT-3 \citep{brown2020language}, a gigantic pre-trained language model. Specifically, we prompt GPT-3 to synthesize an initial set of utterances with intent descriptions and few-shot examples~\citep{kumar-etal-2020-data, yoo-etal-2021-gpt3mix-leveraging}, shown in Figure \ref{fig2}. Then, to scale up training data, we transform the synthetic utterances into templates by replacing slot values with placeholders. The templates are filled with various sampled values, such as task names, to get actual training utterances. Moreover, linguistic rules and neural paraphrase models are used to create mixed-intent utterances and improve data diversity. At last, we inject user noise, such as filler words, to enhance the robustness of our intent recognition module. 

\textbf{Task Name Extraction.} \label{taskname}Given a task request, this module extracts a task name, which will be used by the search engine to construct the search query. As an example, it extracts \textit{``wash a car''} from the home improvement task \textit{``How to wash a car?''} and \textit{``bubble tea''} from the cooking task \textit{``Search bubble tea recipe for me.''} We formulate task name extraction as a span extraction task and fine-tune a BERT-based model \citep{devlin-etal-2019-bert} on our synthesized task requests.

\textbf{Task Domain Classification.} \label{domain}
We train another BERT-based binary classifier on our synthesized task requests to distinguish home improvement and cooking tasks. This enables TacoBot to offer different dialogue experiences for the two different domains. 


\subsection{Safety Check}
\label{safety}
TacoBot ensures safe conversations by (1) \textbf{Profanity Check}, where we use the offensive speech classifier in CoBot \citep{abs-1811-12900} to check both user utterances and system responses in each turn. For offensive user inputs, our system redirects users to their ongoing tasks. We also remove potentially offensive sentences from the final response if detected. (2) \textbf{Task Safety Check.} Following the competition rules, TacoBot politely rejects two kinds of inappropriate task requests: (a) \textit{dangerous tasks}, where users and their properties may get hurt, and (b) \textit{professional tasks}, especially those involving legal, medical, and financial knowledge. To detect these inappropriate task requests, we perform rule-based matching against a keyword blacklist provided by Amazon and a set of over $10,000$ Wikihow article titles annotated by our team. Once detecting (a) \textit{dangerous tasks}, our system directly ends the session; or (b) \textit{professional tasks}, it redirects the user to other appropriate tasks. 

%% file: DM.tex
\begin{figure}[b]
    \centering
    \includegraphics[width=\columnwidth]{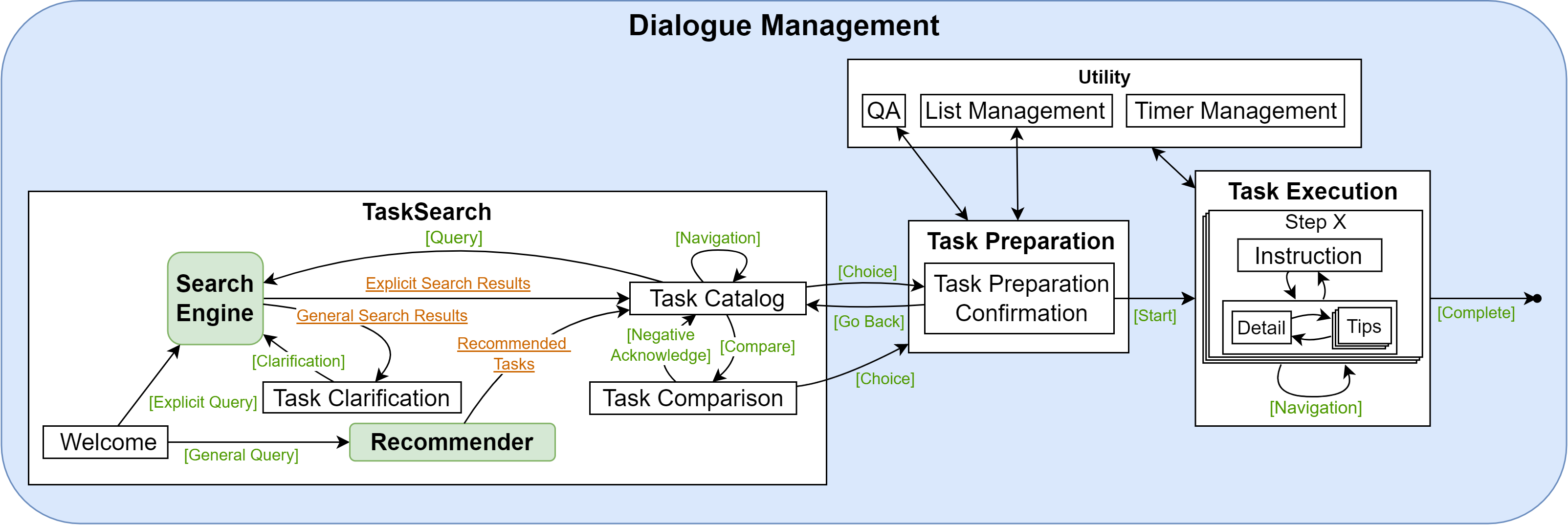}
    \caption{Simplified View of TacoBot's Dialogue Management Module. White boxes represent dialogue states and green boxes represent supporting modules}. Bidirectional edges represent reflexive transitions. \textcolor{LimeGreen}{Green text} represents user intent and \textcolor{Orange}{orange text} represents search engine output. 
    \label{fig_sm}
\end{figure}

\section{Dialogue Management}
\label{dm}

Due to the goal-oriented nature of dialogues in in the Alexa Prize TaskBot challenge,
we divide the TaskBot dialogue experience into three phases supported by a utility module and design a hierarchical finite state machine for DM (Figure \ref{fig_sm}). Each phase has multiple fine-grained dialogue states. 
\begin{itemize}
    \item 
    \textbf{Task Search Phase.} Task Search is the first phase in TacoBot, where users try to find a DIY task or recipe. With a specific idea in mind, users can directly issue a query and get search results from our backend search engine. Alternatively, they can ask TacoBot to recommend interesting tasks. Either way, TacoBot can present and compare candidate tasks retrieved by the search engine to help users make a choice. One unique design of TacoBot is that, when searching for recipes, TacoBot actively asks clarification questions regarding diet constraints or cuisine, such as \textit{nuts-free}, \textit{vegetarian}, \textit{Chinese food}, and \textit{Mexican food}, to present more accurate search results. The users then enter the Task Preparation phase once they pick an option.
    
    \item
    \textbf{Task Preparation Phase.} Users can enter Task Preparation if and only if they have selected a candidate task. In this phase, users review detailed information of the chosen task and decide if they want to proceed. Users also have access to list management features and the QA module at this phase. If users change their mind, they can go back to Task Search and find another task. Otherwise, they can commit to the task and enter Task Execution. 
    
    \item
    \textbf{Task Execution Phase.} After entering Task Execution, users are not allowed to change phases, per the competition rule. In this phase, TacoBot walks the user through the instructions step-by-step to help them complete a task with the assistance of the utility module.  Each step of a task has its own state in this phase. We further break long WikiHow steps into shorter \textit{Instruction}, \textit{Detail}, and \textit{Tips} to make it easier for the user to digest. 
    
    \item
    \textbf{Utility Module.} TacoBot users can access various utility features, including question answering, help information, as well as Alexa list and timer management through the Utility Module. The utility features are invoked by intents and do not affect the states in the three TacoBot phases. 
\end{itemize}

Upon user input, the dialogue manager performs state transition and choose response generators (Section~\ref{nlg}) accordingly.
We impose strict conditions for transitions between phases. For example, a specific task must be selected when transitioning from the \textbf{Task Search} phase to the \textbf{Task Preparation} phase. 
Within each phase, we assign a dialogue state for each possible conversation turn hierarchically.
This hierarchical dialogue state design allows us to define flexible transitions at different levels. 
For example, users can access the utility module from any state in the \textbf{Task Execution} phase.
We also maintain a dialogue state history stack to allow users to go back to previous states easily.
User intents that do not make valid transitions have no impact on dialogue states. Instead, they trigger contextualized help information that guides users to proceed with their dialogues. 
As a result, TacoBot provides stable yet flexible dialogue experiences.


%% file: TaskSearchRerank.tex
\section{Search Engine}
\label{search}

The task search quality is essential to user experience. We first build a search engine for cooking (from WholeFoodsMarket) and home improvement (from WikiHow) tasks based on elastic search, and further improve the search results with query expansion
. A neural re-ranking module is developed to rerank search results for home improvement tasks.

\subsection{Query Expansion}
One major flaw of using elastic search is that it overemphasizes lexical similarity between the input query and task titles, leading to semantically mismatched search results in many cases. While task titles are usually long and comprehensive, real user requests tend to be short, long-tailed paraphrases of them, making it even more difficult for searching. We develop a query expansion technique to alleviate the impact of such mismatch and improve search result relevance. 
Query expansion expands task names in user queries by adding related words to improve the recall of search results, including lemmatized verb, nouns, and decomposed compound nouns. For example, the expanded query for \textit{``How to remove spraypaint''} is [``how'', ``to'', ``remove'', ``spraypaint'', ``spray'', ``paint''].




\subsection{Neural Re-ranking}


Since our system shows three search results at a time, hit rate of Top-3 and Top-6 is very important to improve the precision of search results shown to the users. Particularly, for home improvement tasks, the lexical mismatch between user task request and task titles is noticeable. To further improve search performance, we build a neural re-ranking model that brings semantically related search results to the front of the list of candidate tasks. Based on BERT-based model, the re-ranker takes a task request and retrieved task titles as input and assigns a score for each task. We adopt a weakly-supervised list-wise ranking loss \citep{cao2007learning} for training, where one positive and $n$ negative samples ($n = 9$) are used in each step. 

To avoid the cost of human annotation, the re-ranker is trained on synthesized task queries via GPT-3 query simulation, as described in Section~\ref{nlu}. We propose to collect weak supervision signals from Google's search engine. 
Specifically, we limit the search space to Wikihow when searching for a home improvement task by appending the suffix, ``\textit{site:wikihow.com}'', to the task request. The top-3 returned search results are used as positive labels for that task request. We collect hard negative samples by querying elastic search with the same requests and finding any results that do not overlap with Google's top-3 ones. 

%% file: NLG.tex
\section{Response Generation}
\label{nlg}
Our response generation module mainly leverages handcrafted conditional rules to fill slots with retrieved data and glue templated segments together. 
For task-related questions, we build a neural question answering model to detect their relevance and extract their answers from contexts. 

\subsection{Template-Based Generation}
\label{template}
TacoBot stores the response templates as segments of phrases and sentences. The templated segments are carefully curated by native speakers and have several pre-written paraphrases. At composition time, they are selected randomly to generate diverse, human-like responses. We organize the templates and their composition rules according to the high-level states in our hierarchical finite-state machine. 

When presenting structured data, TacoBot uses sentence-level templates and substitutes named slots with their data values. For instance, our task preparation response provides descriptive information about the candidate task. To generate this response, the module selects and fills the template based on attributes like ratings, popularity, and estimated time. When presenting unstructured texts, TacoBot uses regular expressions to segment them into short, conversation-friendly pieces and combines them with phrasal or sentence-level templates. During task execution, for example, we index each step with prefixes and append possible commands for users to try in their response. 

\textbf{Alexa List and Timer Service.} CoBot also provides interfaces to two kinds of useful Alexa service, list and timer. We recognize that they are essential for users to complete their tasks. Therefore, when users intend to invoke these two kinds of service, we enable TacoBot to generate templated acknowledgements and fill in slots describing its action, similar to how it presents structured data. 


\subsection{Question Answering}
\label{qa}
Question answering (QA) is a crucial functionality of task-oriented dialogue systems. In addition to the open-domain EVI system provided by Amazon, TacoBot's QA module has an in-context machine reading comprehension (MRC) module for context-dependent QA, a frequently-asked questions (FAQs) retrieval module for DIY tasks and a rule-based ingredient and substitute QA module for cooking tasks. On top of these QA modules, we build a question type classifier to decide which QA module to call given a question from the user.

\subsubsection{Question Type Classifier} We build a question type classifier that classifies a user question into 5 types (\texttt{MRC, FAQ, Factual, Ingredient, Substitute}) under a cooking task and 3 types under a DIY task (\texttt{MRC, FAQ, Factual}). 
\texttt{Factual} questions are taken by the EVI system and other questions are handled by corresponding QA modules (see below). 
To better differentiate different types of questions, we concatenate the instruction of the current step (if available) as context with the input question and feed the sequence into a \texttt{Roberta-base} classifier. We sample 5,000 questions for each type (see examples in Table \ref{tbl:eg_diff_ques}) in our training set. 


\begin{table}[h]
\small
\centering
\begin{tabular}{l|l|l}
\hline
Question Type & Example & Context \\ \hline
MRC & Use what tool to blend? & \multirow{5}{*}{\begin{tabular}[c]{@{}l@{}}add a 14-ounce can of sweetened \\ condensed milk, unsweetened natural \\ cocoa powder... use a whisk to blend \\ the ingredients until they’re completely \\ mixed, and set the bowl aside. it’s\\ normal for the mixture to become \\ very thick as you mix it.\end{tabular}} \\ \cline{1-2}
Factual & How much is 14 ounce in gram? &  \\ \cline{1-2}
FAQ & \begin{tabular}[c]{@{}l@{}}How should I know the ingredients \\ are completely mixed?\end{tabular} &  \\ \cline{1-2}
Substitute & \begin{tabular}[c]{@{}l@{}}I don't have condensed milk, \\ can I use something else?\end{tabular} &  \\ \cline{1-2}
Ingredient & How much cocoa powder do I need? &  \\ \hline
\end{tabular}
\caption{Examples of different types of questions.}
\label{tbl:eg_diff_ques}
\end{table}

\subsubsection{Context-Dependent QA}
We started with UnifiedQA \citep{emnlp/KhashabiMKSTCH20}, a pre-trained language model for robust cross-domain QA, and observed two issues in real user conversations: (1) it often fails to detect \textit{unanswerable} questions; and (2) it sometimes generates unrelated hallucinated answers. For example, when the user asks \textit{``Where can I buy peanut seeds and soil?''} during executing ``\textit{grow nuts}'', the model generates \textit{``peanuts are sold at my local stores''} as the answer. To solve these issues, we annotate an in-context QA dataset and fine-tune an extractive QA model.

\textbf{Data Annotation.} Following the annotation protocal of SQuAD \citep{emnlp/RajpurkarZLL16,acl/RajpurkarJL18}, we first sample $1832$ paragraphs\footnote{A paragraph corresponds to\nop{usually} a step in a WikiHow article.} as context from WikiHow articles for cooking and home improvement tasks. Then, we ask 15 graduate students to create up to 3 question-answer pairs for each paragraph. The annotated answers are either sentences in the context for answerable questions or a special `[No Answer]' token for unanswerable questions. 
In total, we obtain $5183$ QA pairs including 752 \textit{unanswerable} questions (Table~\ref{tbl:statistics}). 
\begin{table}[!h]
\centering
\begin{tabular}{c|ccc}
\hline
\multirow{2}{*}{\# of samples} & context & answerable QA & unanswerable QA \\
 & 1832 & 4431 & 752 \\ \hline
\multirow{2}{*}{\begin{tabular}[c]{@{}c@{}}avg len\\ (\# of words)\end{tabular}} & context & question & answer \\
 & 93.9 & 9.76 & 17.9 \\ \hline
\end{tabular}
\caption{Statistics of our annotated WikiHow QA dataset.}
\label{tbl:statistics}
\end{table}

\textbf{Model Fine-tuning.} To avoid hallucination, we use {Roberta-base} \citep{corr/abs-1907-11692} to develop an extractive QA model in two stages. 
We pre-train our model on SQuAD2.0 \citep{acl/RajpurkarJL18} before fine-tuning on our annotated QA dataset.
Taking the observation that user may ask questions about previously shown steps, we augment the context by concatenating the current step with the previous $n$ steps ($n=2$) for training and inference.



\subsubsection{Context-Independent QA}


\textbf{FAQ Module.}
Frequently-asked questions (or Expert QAs) are important knowledge sources in WikiHow articles. Since the questions are raised by the real users and answers are from human experts, such FAQ pairs can reliably answer similar questions from TaskBot users. We collect QA pairs from the \textit{Community Q\&A} section of WikiHow articles and use them as our FAQ knowledge source. TacoBot's FAQ QA module is a retrieval module based on the cosine similarity between question embeddings, which are produced by a sentence-BERT \citep{rg19}\footnote{\url{https://huggingface.co/sentence-transformers/all-mpnet-base-v2}} encoder. At inference time, we use a similarity threshold of $t=0.75$.

\textbf{Ingredient and Substitute QA Module.}
To support questions regarding ingredients in cooking tasks, we extract the ingredient mentioned by users through a high-recall string matching mechanism against the list of ingredients in the chosen recipe. When users don't have a particular ingredient, TacoBot could try to offer substitution suggestions if possible. To that end, we collect a substitution data set which covers 200 frequently used ingredients.


%% file: UE_short.tex
\section{User Engagement}
\label{ue}

While a taskbot is primarily a task-oriented system, we observe early in the competition that the bot was frequently used as an interesting novelty to experiment with.
In a study of randomly sampled conversations in the quarterfinals, we find that of the 74.5\% conversations in which a task was initiated, approximately 44\% of the tasks were sample tasks suggested in the welcome prompt.
Further, we find that in about 10\% of the conversations, users attempted to initiate non-task related ``chatting'' with the bot, frequently asking personality-related questions like  \textit{``What's your favorite food?"}. 
These findings lead us to the conclusions that (1) recommending interesting tasks is likely to improve user satisfaction, and (2)  more lively responses can make interactions more enjoyable. Thus, we try to improve user engagement by upgrading our example tasks to curated ``favorite'' tasks and infusing personality traits into TacoBot's responses.

\textbf{TacoBot's Favorites.}
We implement a ``favorites'' feature where we hand-select tasks to recommend to the user. These tasks are selected based on the quality of the content, seasonality, and customer ratings, but also their relevance to a set of predetermined personality traits and hobbies we decided upon for our bot. For example, we decide that our TacoBot's interests include plants, animals, crafts, and home decor and therefore most of the ``favorite'' tasks are on these topics. These tasks effectively boost the average user rating for conversations in which users accessed the favorites feature in comparison to conversations in which they did not. Additionally, we plan to further introduce hand-crafted featured articles, which are Wikihow articles rewritten by our team members with extra fun facts, tips and more interesting language. We expect these featured articles to provide users with more engaging conversations and pique their interest in topics unique to our bot. 

\textbf{Improving Rapport.} In order to make interactions more engaging and increase user trust, we implement three features: redesigning response templates to be more varied and personable, improving the context awareness of responses by making them dependent upon the user's time of day, and creating an attractive visual interface for users with screens.




%% file: test-suite.tex
\section{Automated Test Suite}
\label{test}


Automating tests for task-oriented dialogue systems is especially difficult because TaskBots (1) deal with diverse inputs and outputs and (2) rely on third-party services. Nonetheless, the complexity of TacoBot raises our awareness of testing. Therefore, we write an end-to-end test suite from scratch. Our end-to-end tests allow us to identify issues in the system efficiently and fix them before deployment, leading to stronger ratings in the semifinals.



\textbf{Keyword-Based Testing.}
Typical tests compare an expected result with an actual result and fail if they are different. 
Because TacoBot has diverse responses for each dialogue state, exact string matching is too rigid for testing.
Instead, we tested for the presence of keywords that all appropriate responses must contain. 
Meanwhile, our tests forbid a different set of keywords that any fallback responses would have. 
For example, we ensure that the response to ``tell me your favorites'' includes the words ``recipe,'' ``task,'' and ``favorite'' and does \textbf{not} include ``sorry'' \textbf{nor} ``don't understand.''

\textbf{Taco Monitor.}
We also build an internal website, Taco Monitor, to review all user interactions and identify any issue in near real time. For important problems we want to eliminate in future dialogues, we add the entire conversation into our test suite. For instance, we observed that when users said ``cancel'', TacoBot repeated its last response, instead of reminding the user of unable to cancel per the competition rule and prompting help information. By adding one of those conversations as a test case, we prevented the bug from happening again. 

%% file: results.tex
\section{Results and Analysis}
\label{exp}

\subsection{System-level Performance}

\begin{figure}[h]
  \centering
  \begin{minipage}[b]{0.47\textwidth}
    \includegraphics[width=\textwidth]{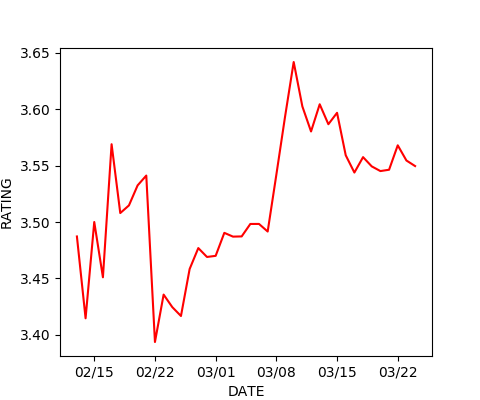}
  \end{minipage}
  \hfill
  \begin{minipage}[b]{0.52\textwidth}
    \includegraphics[width=\textwidth]{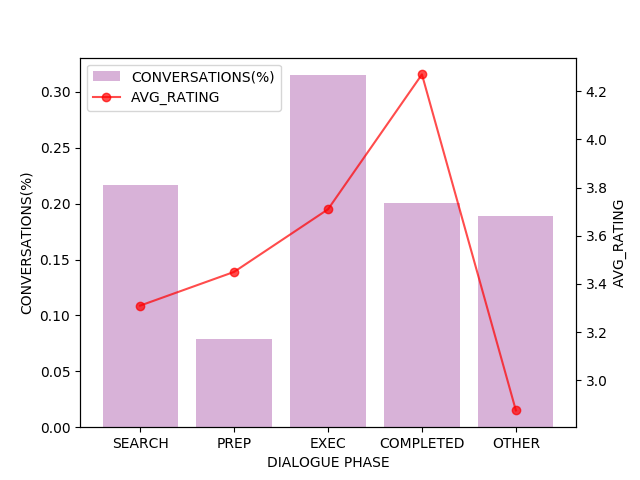}
  \end{minipage}  \caption{User Ratings from 02/07/22 to 03/24/22. Left: TacoBot's average rating during the semifinals. Right: The number and average rating of conversations by dialogue phases (Section \ref{dm}) they end in. }
  \label{fig3}
\end{figure}

Based on our analysis of the sampled quarterfinal conversations (Section \ref{ue}), we hypothesize that user ratings positively correlate with their progress on the tasks. Thus, we prioritize improving two functionalities of TacoBot, finding relevant search results and offering engaging recommendations, to increase the number of users who start their tasks. As a result, we observe an increase of TacoBot's average rating and a correlation between user ratings and the phases they end in, shown on the left and right of Figure \ref{fig3}, respectively.

Specifically, $21.65\%$ conversations end in the Task Search phase with an average rating of $3.31$, mostly due to irrelevant search results. In contrast, if TacoBot presents satisfying candidate tasks, only $7.87\%$ conversations (average rating: $3.45$) are suspended in the Task Preparation phase, and users of $51.58\%$ conversations proceed to start their tasks. Among them, users of $31.50\%$ conversations stop in the middle of a task with an average rating of $3.71$, and $2.36\%$ resumed and completed the tasks in a new dialogue session. $17.72\%$ more conversations record task completion within one session. The average rating of all $20.08\%$ conversations with tasks completed is $4.27$. However, the remaining $18.90\%$ conversations only have an average rating of $2.88$. Most of these conversations include serious exceptions, such as no search results returned, inappropriate task requests, and unexpected user behaviors like chatting. 

\subsection{Module-level Performance}
\label{mod-perf}

\textbf{Natural Language Understanding.} We compare our three neural models for NLU (intent recognition, task name extraction, and domain classification) with the built-in interaction model of Alexa Skill Kit (ASK). The evaluation dataset contains utterances collected from real user conversations between 09/16/21 and 12/07/21.
We measure the performance on intent recognition and domain classification with accuracy and that on task name extraction with exact match (EM) and Span F1 (Table~\ref{tab:NLU}). In summary, our neural models consistently outperform ASK on all three tasks.

\begin{table}[h]
    \centering
    \begin{tabular}{lM{2cm}M{2cm}M{2cm}M{2cm}}
    \toprule
    &Intent Recognition&\multicolumn{2}{M{4cm}}{Task Name Extraction}&Domain Classification\\\cmidrule(lr){2-2}\cmidrule(lr){3-4}\cmidrule(lr){5-5}
    &Accuracy&EM&Span F1&Accuracy\\
    \midrule
    ASK & 63.5&30.2&58.1&61.1 \\
    Neural Models  & \textbf{78.7}&\textbf{96.3}&\textbf{98.6}&\textbf{93.3}\\ 
    \bottomrule
    \end{tabular}
    \caption{Results for Our Neural NLU Models.}
    \label{tab:NLU}
\end{table}

\textbf{Search Engine.} To measure our search engine's performance, we use real user queries and GPT-3 simulated ones to construct an evaluation dataset. For each query, we retrieve results from Google search and label all relevant candidates as positive. We collected 700 test queries with at least one positive result per query and use hit rate in top-$k$ (HIT-$k$) to evaluate our search engine.

As shown in Table~\ref{tab:reranking}, query expansion improves HIT-3 and HIT-6 by around $7\%$, compared to the baseline using unmodified queries. We also calculate HIT-25 with query expansion to estimate an upper bound for neural re-ranking. Augmented with the re-ranker, our search engine can closely approximate the upper bound with high HIT-3 ($76.9\%$) and HIT-6 ($78.9\%$). Particularly, the overall HIT-6 is only $2.8\%$ lower than the upper bound (HIT-25), indicating most of the time our re-ranker can get at least one positive in top-6 as long as there is one in the initial top-25 list. Note that we use the DistilBERT as the backbone of the re-ranker to reduce inference-time latency.

Also, we split the test set based on whether the initial search results with only query expansion contain at least one positive in top-$k$ (Easy-$k$) or not (Hard-$k$). This split shows that our re-ranker can largely improve the search performance on the hard sets with a limited loss on the easy sets.


\begin{table}[h]
\centering
\begin{tabular}{@{}lcccccc@{}}
\toprule
                    & Overall & Easy-3 & Hard-3 & Overall & Easy-6 & Hard-6 \\ \midrule
Upper Bound {(HIT-25)} & 81.7   & 100    & 51.9     &81.7         & 100    & 37.6     \\ \midrule 
                    & \multicolumn{3}{c}{HIT-3}    & \multicolumn{3}{c}{HIT-6}    \\ \cmidrule(lr){2-4}\cmidrule(lr){5-7} 
Original Queries       & 55.7   & -    & -        & 63.4   & -    & -        \\
+ Expansion       & 62.0   & 100    & 0         & 70.7   & 100    & 0         \\
+ Expansion + Re-ranking        & \textbf{76.9}   & 96.3  & \textbf{45.1}     & \textbf{78.9}  & 97.8  & \textbf{33.2}     \\ \bottomrule
\end{tabular}
\caption{Results of Our Query Expansion and Neural Re-ranking. Easy-$k$ represents the set of samples where at least one positive is found in the initial top-$k$, while Hard-$k$ is the set with no positive found in the initial top-$k$. {Upper bound (HIT-25) and initial top-$k$ are based on query expansion.} \nop{The numbers of upper bound (HIT-25) and the initial top-$k$ are from the query expansion.}}
\label{tab:reranking}
\end{table}


\textbf{Question Answering.} We tested the two neural models in our QA module, question type classifier and machine reading comprehension model, with manually annotated test sets. For the question type classier, we evaluated it with 500 questions and observed an overall accuracy of 94\%, which indicates the classifier is proficient at deciding the question types.


For the machine reading comprehension (extractive QA) model, we evaluate it on two test sets: (1) our own annotated test set and (2) questions in real TacoBot conversations (12/10/21 - 02/08/22) with answers annotated by a team member.
As shown in Table \ref{tbl:qa_performance}, UnifiedQA and Roberta achieve comparable performance on \textit{answerable} questions. However, UnifiedQA performs much worse in identifying unanswerable questions on both test sets. This is reasonable as UnifiedQA was pretrained on data sources which do not contain many unanswerable questions. To provide better user experience, it is more advisable to give no answer than some random answer. Thus, we adopt Roberta, an extractive model, as the backbone of our QA module.

\begin{table}[h]
\centering
\resizebox{\linewidth}{!}{
\begin{tabular}{l|cc|cc}
\hline
 & \multicolumn{2}{c|}{Test set (team created)} & \multicolumn{2}{c}{Test set (from real user)} \\ \cline{2-5} 
 & Answerable (436) & Unanswerable (84) & Answerable & Unanswerable \\ \hline
UnifiedQA & 39.3 & 5.9 & 10.3 & 22.4 \\
+finetuning & \textbf{69.9} & 49.2 & 72.6 & 43.2 \\ \hline
Roberta & 40.1 & 48.8 & 40.3 & 88.3 \\
{+finetuning} & 69.7 & \textbf{69.1} & \textbf{72.3} & \textbf{88.9} \\ \hline
\end{tabular}
}
\caption{Accuracy (Exact Match) of QA Models. Number in parenthesis means test set size. }
\label{tbl:qa_performance}
\end{table}

%% file: conclu-v2.tex
\section{Discussion and Future Work}

Until the semifinals, TacoBot has achieved a $42.0\%$ relative increase in average rating, from $2.50/5.0$ to $3.55/5.0$. In the semifinals, we validated our hypothesis that user satisfaction correlates with accurate search results and attractive recommendations. Nevertheless, TacoBot can be further improved on multiple fronts. 

On the one hand, better conversational search and recommendation are required to secure basic user satisfaction. After the semifinals, we introduced a single-turn clarification for recipe searches, but it is still less flexible and limited to diet constraints and cuisines. Instead of increasing the number of clarification questions, it would be interesting yet challenging to enable TacoBot to have a mixed-initiative chitchat around the user's initial query while extracting search constraints and user preferences. On the other hand, we envision TacoBot to be more engaging and accessible. For example, incorporating knowledge-grounded generation models to augment the curated templates would potentially make the responses more conversational and diverse. Also, adding more commands such as volume and pace control could further enhance TacoBot's accessibility. 

Finally, the evaluation of TaskBots is a grand open challenge. While Section \ref{mod-perf} provides module-level performances on our constructed datasets, it is hard to measure the impact of each module on the overall system in real conversations. So far, the performance of the overall system is only evaluated by one single user rating per dialogue, making it harder to measure the correlation between each component and user satisfaction. Moreover, as an evaluation metric, average user rating requires a sufficiently large scale to reflect the TaskBots' overall performance and to conduct A/B tests. Therefore, we believe it is worthwhile exploring other novel methods (automatic and interactive) to assess the quality of TaskBots. 